\pdfoutput=1
\def\year{2020}\relax

\documentclass[letterpaper]{article} 
\usepackage{aaai20}  
\usepackage{times}  
\usepackage{helvet} 
\usepackage{courier}  
\usepackage[hyphens]{url}  
\usepackage{graphicx} 
\usepackage{lineno,hyperref,amsmath,amssymb}
\usepackage{booktabs}
\usepackage{multirow}
\usepackage[normalem]{ulem}
\useunder{\uline}{\ul}{}
\usepackage[table,xcdraw]{xcolor}
\modulolinenumbers[5]
\usepackage{graphicx}  
\usepackage{amsmath}
\usepackage{amsfonts}
\usepackage{amsthm}
\usepackage{booktabs}

\usepackage{dsfont}
\usepackage{multirow}
\usepackage{enumerate}
\usepackage{pifont}
\usepackage{subcaption}

\newcommand{\citet}[1]{\citeauthor{#1}~\shortcite{#1}}
\newcommand{\citep}{\cite}

\nocopyright
\title{QuatDE: Dynamic Quaternion Embedding for Knowledge Graph Completion}
\author{Haipeng Gao, Kun Yang, Yuxue Yang, Rufai Yusuf Zakari, Jim Wilson Owusu, Ke Qin\thanks{Corresponding author.}\\
University of Electronic Science and Technology of China\\ 
\{gaohp, yangkun, yuxueyang\}@std.uestc.edu.cn\\
\{rufai6, owilsonjim\}@gmail.com\\
\{qinke\}@uestc.edu.cn
}
 
\begin{document}

\maketitle

\begin{abstract}
Knowledge graph embedding has been an active research topic for knowledge base completion (KGC), with progressive improvement from the initial TransE, TransH, RotatE et al to the current state-of-the-art QuatE. However, QuatE ignores the multi-faceted nature of the entity and the complexity of the relation, only using rigorous operation on quaternion space to capture the interaction between entitiy pair and relation, leaving opportunities for better knowledge representation which will finally help KGC. In this paper, we propose a novel model, QuatDE, with a dynamic mapping strategy to explicitly capture the variety of relational patterns and separate different semantic information of the entity, using transition vectors to adjust the point position of the entity embedding vectors in the quaternion space via Hamilton product, enhancing the feature interaction capability between elements of the triplet.  Experiment results show QuatDE achieves state-of-the-art performance on three well-established knowledge graph completion benchmarks. In particular, the MR evaluation has relatively increased by 26\% on WN18 and 15\% on WN18RR, which proves the generalization of QuatDE.
\end{abstract}

\section{Introduction}

Billions of facts in the world can be stored in the Knowledge Graphs (KGs) with triples succinctly, and each triple $(h,r,t)$ consist of two nodes and a directed edge between them. KGs such as Freebase\cite{bollacker2008freebase}, YAGO\cite{suchanek2007yago} and DBpedia\cite{lehmann2015dbpedia} are useful in many AI applications, such as question answering (QA)\cite{cui2019kbqa}, recommended system\cite{wang2018dkn}, relation extraction\cite{wang2020direction}, etc. Even though the KGs have been studied for many years, they still suffer from incompleteness, which makes their downstream assignments more challenging. As a result, researchers have put more focus on knowledge graph completion (KGC) task, which dedicates to predict missing links between nodes. In other words, KGC infers the implicit triples based on the true triplets that exist in the KGs. For example, if the triple (\textbf{Bill Clinton}, \textbf{\textit{Friendship}}, \textbf{Seven Spielberg}) is correct, i.e. \textbf{Bill Clinton} has a \textbf{\textit{friendship}} with \textbf{Seven Spielberg}, we can infer that \textbf{Seven Spielberg} is a \textbf{\textit{friend}} of \textbf{Bill Clinton} equally, i.e. (\textbf{Seven Spielberg}, \textbf{\textit{Friendship}}, \textbf{Bill Clinton}).

\begin{figure}[h]
    \centering
    \includegraphics[width=0.9\columnwidth,height=!]{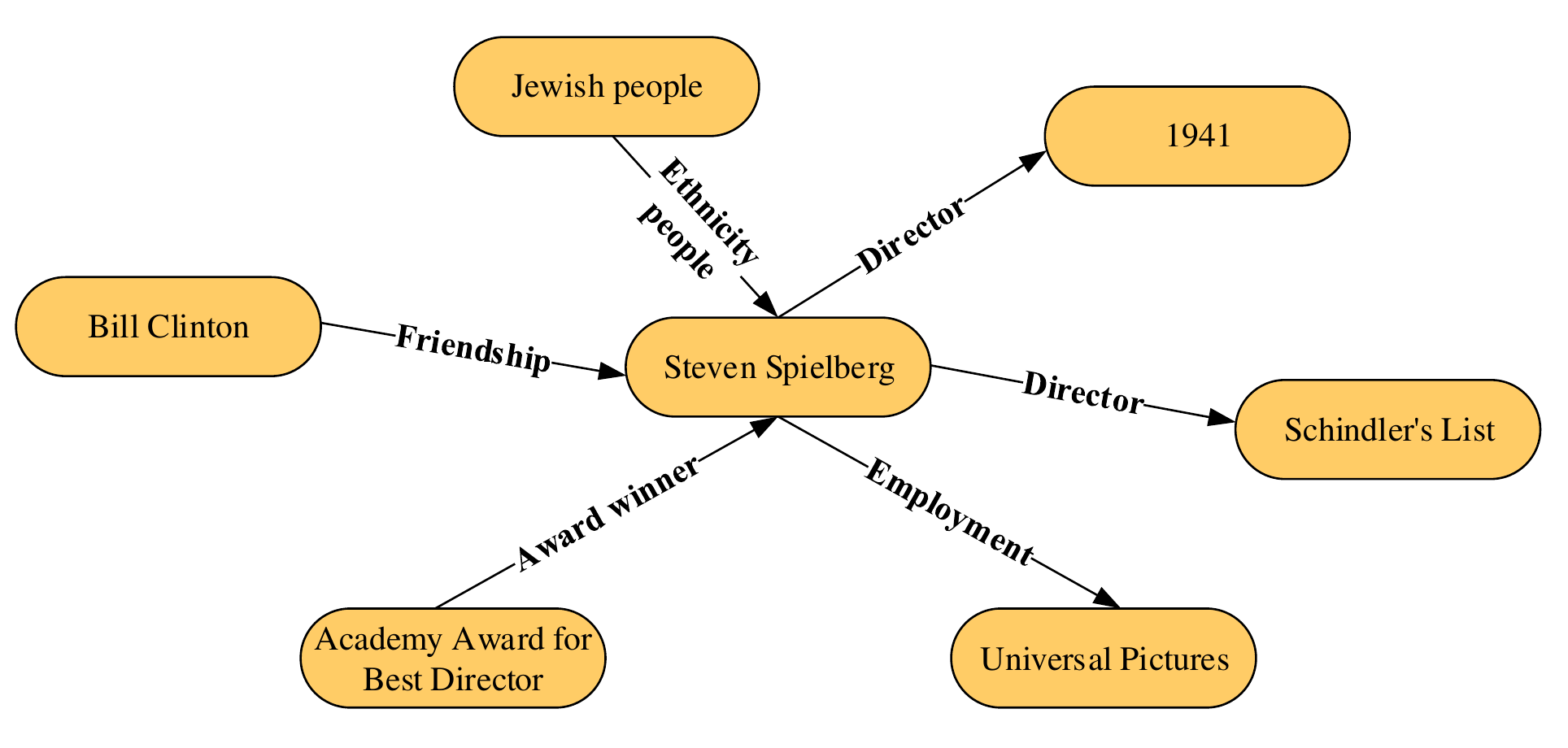}
    \caption{\textbf{subgraph of knowledge graph in FB15K-237. Entities are represented as golden blocks and directed links are represented as black arrows}}
    \label{img1}
\end{figure}

Conventional approaches for KGC have achieved substantial improvement via embedding entities and relations into low-dimensional continuous space, such as TransE\cite{bordes2013translating}, TransH\cite{wang2014knowledge}, TransD\cite{ji2015knowledge}, TransR\cite{lin2015learning}, etc. Instead of using a real-valued space, ComplEx\cite{trouillon2016complex}, RotatE\cite{sun2019rotate} project entities and relations into a complex space preforming the strong representation ability. Meanwhile, QuatE\cite{zhang2019quaternion}, Rotate3D\cite{gao2020rotate3d} show the rich feature interaction capacity between triple with Hamilton product in quaternion space, obtaining state-of-the-art (SOTA) link prediction results.

\begin{figure*}[htbp]  
    \begin{minipage}[t]{0.5\linewidth}  
        \centering  
        \includegraphics[width=0.9\columnwidth,height=!]{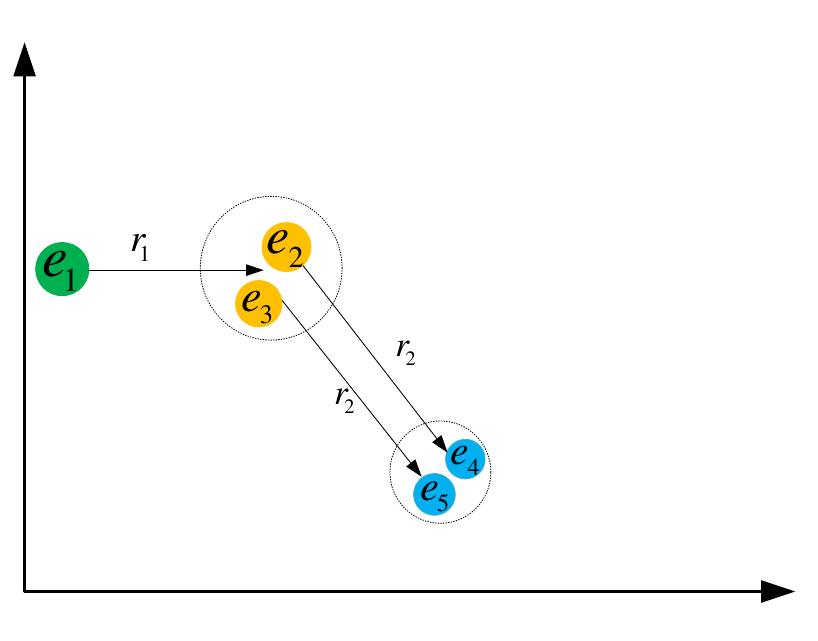}  
        \caption*{(a) QuatE}  
    \end{minipage}%
    \begin{minipage}[t]{0.5\linewidth}  
        \centering  
        \includegraphics[width=0.9\columnwidth,height=!]{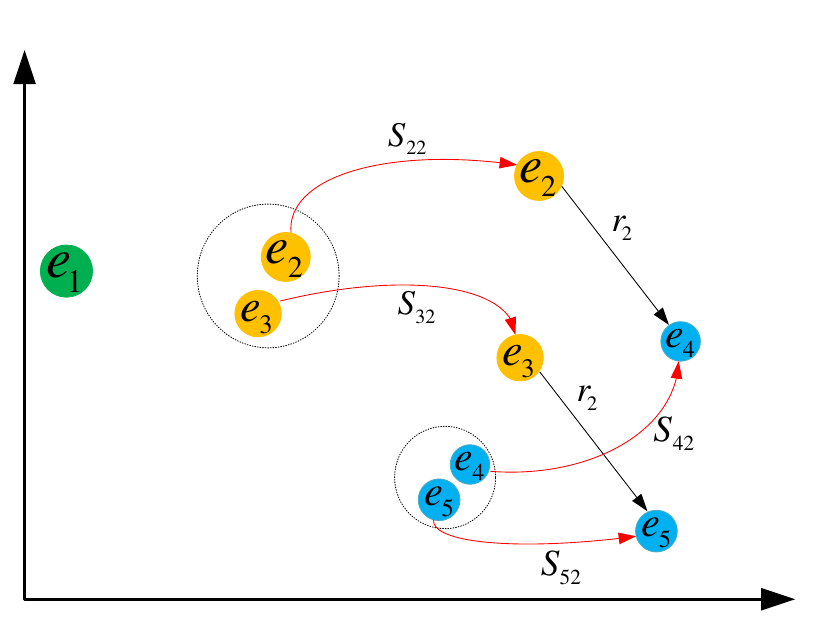}  
        \caption*{(b) QuatDE}  
    \end{minipage}
    \caption{\textbf{Simple illustration of QuatDE, Entities ($e_m$) and directed links ($r_n$) are represented as colored circles and black arrows, respectively. The red curves ($S_{mn}$) symbolize the dynamic mapping strategy which are determined by the elements in different triples. We draw similar entities into one dotted circle.}}  
    \label{img2}
\end{figure*}

Now, although most models surrounding the KGC task are powerful enough, handling complex relation patterns are still a major challenge. QuatE utilizes quaternion embeddings to represent entities and relations. Relations are modelled as rotations enjoying the highly expressive ability. However, QuatE has two main problems: 1) failing to dig deep information, capturing the ability of representation and feature interaction between entities and relations are relatively weak, because it only relies on the rigorous rotation calculation of three embedding vectors; 2) Various relation patterns including one-to-many, many-to-one, and many-to-many are not to be considered. Figure \ref{img2} (a) shows the positions of entities and relations in quaternion space determined by QuatE are absolute rather than continuously changing, resulting the distances of similar entities (such as 1941 and Schindler’s List, directed by Steven Spielberg) in the space are close. It’s detrimental to predict the fact who played in 1941 and Schindler’s list respectively.

To address this issue, we propose a new model QuatDE. The basic idea is illustrated in Figure \ref{img2} (b), we consider the deep and multifaceted meaning of the entity. The same or similar entities in different triplets are represented by distinct vector representations, dynamically determined by specific relations and positions (head or tail). So, based on QuatE, we introduce a dynamic mapping strategy ($S_{mn}$) consisting of three additional quaternion vectors (defined as subject transfer vector, object transfer vector and relation transfer vector). It used to adaptively adjust the position and separates different semantic information of the entity. Meanwhile, the model gains full expressive ability.

\paragraph{ \textbf{Our Contributions}}In summary, our proposed model has the following contributions:

\begin{itemize}

\item We propose a novel knowledge graph embedding method in the quaternion space, using a dynamic mapping method (strongly related to entities and relations) to explicitly enhance the interaction among triplets, modeling the diversity of relations and multiple semantics of the entities.
\item Our model has multiple-level improvements in embedding dimensions and has strong generalization capabilities (MR). To be precise, MR and Hit@10 of our model are still better than the QuatE model when the embedding dimension is 50 in WN18 (one-sixth compared with QuatE).
\item QuatDE is an excellent result of combining QuatE and quaternion-valued neural network and we expound on the superiority of QuatDE from the perspective of QNN.
\item Our method is extended to standard benchmark datasets: FB15K-237, WN18 and WN18RR. Experiment results prove that our method is superior to the previous methods, and our code can be available on GitHub: \url{https://github.com/hopkin-ghp/QuatDE}.

\end{itemize}

\section{State of the art}

In this section, we will roughly describe the related work in two parts: translational distance models and semantic matching models. Note that semantic matching models exploit similarity-based scoring functions including bilinear models and neural network based models. Then we will discuss the connection between our approach and others.

\subsection{translational distance models}

The translational distance models utilize distance-based cost function and are efficient with low computational cost. We will describe this type of models with a unified formula, the scoring function is designed as follows:

\begin{equation}
f_{r}(h, t)=|| S_{r}(\boldsymbol{h})+\boldsymbol{r}-S_{r}(\boldsymbol{t}) \|_{l 1 / l 2}
\end{equation}in which function $f_r (h,t)$ represents the score of triple $(h,r,t)$, and $S_r ( \cdot )$ donates as the linear function, projecting entity embedding vectors in the relation-specific vector space. Generally, such models measure distance from source entities to target entities with $l1$-norm or $l2$-norm.

TransE\cite{abboud2020boxe} is the most primitive but prominent model: $S_r (\boldsymbol{h})=\boldsymbol{h}$, $S_r (\boldsymbol{t})=\boldsymbol{t}$. However, TransE does not do well in dealing with 1-to-N, N-to-1, N-to-N relations. To solve this problem, the variants of above model, TransH, TransD, TransR, etc, consider unequal mapping strategy to project relation embedding in vector or matrix space, or capturing relational interactions\cite{ji2020survey}. TransH\cite{wang2014knowledge} interprets a relation as a hyperplane with a translation operation, thus, original entity embedding vectors are projected into corresponding relation hyperplane. TransH donates the normal vector in the hyperplane as $w_r$:  $S_r (\boldsymbol{e})=\boldsymbol{e}-\boldsymbol{w}_{\boldsymbol{r}}^{\top } \boldsymbol{e} \boldsymbol{w}_{\boldsymbol{r}}$. TransR\cite{lin2015learning} proposes to use a relation-specific projection matrix $M_r$ rather than hyperplane to project the entities embedding vectors into the space: $S_r (\boldsymbol{e})=M_r \boldsymbol{e} $. It can be seen that large-scale parameters will be designed, so training the model is demanding which requires a lot of storage space. TransD\cite{ji2015knowledge} models the relational mapping matrix $M_{re}$ in a more flexible way: $S_r (\boldsymbol{e})=M_{re} \boldsymbol{e}$, where $M_{re}=\boldsymbol{w}_{\boldsymbol{r}} \boldsymbol{w} _{\boldsymbol{e}}^ {\top} + \boldsymbol{I}$. TranSparse\cite{ji2016knowledge} leverages a numerical space to deal with the heterogeneity and imbalance issues of KGs. TransM\cite{fan2014transition} focus on the structure of the knowledge graph via pre-calculating the distinct weight for each training triplet according to its relational mapping property: $f_r (h,t)= -\theta_{r}\|\boldsymbol{h}+\boldsymbol{r}-\boldsymbol{t}\|_{l_1 / l_2}$. TransAP\cite{zhang2020improve} notes that scoring function based on translation can’t deal with the circle structure and hierarchical structure, so it introduces position-aware entity embeddings and attention mechanism to capture different semantic of triples.

\subsection{semantic matching models}
To be precise, translation models only obtain shallow linear characteristics through simple subtraction or multiplication operations. The scoring functions of semantic matching models reflect the confidence of the semantic information of the triples. RESCAL\cite{nickel2011three} represents each relation as a full rank matrix, optimizing a scoring function that computes a bilinear product between head and tail entity embedding vectors and relation matrix. Due to the large number of parameters, the model has overfitting problem. To alleviate the issue, DistMult\cite{yang2014embedding} uses a diagonal matrix for each relation which reduces parameters to a certain extent. Subsequently, ComplEx\cite{trouillon2016complex} extends DistMult to the complex-valued space, and use a trick that head and tail entity embeddings of the same entity are complex conjugates. SimplE\cite{kazemi2018simple} and TuckER\cite{balavzevic2019tucker} build on canonical polyadic (CP) and Tucker decomposition, respectively. TuckER shows that several linear models, RESCAL, DistMult, ComplEx, SimplE, are special cases of TuckER.

To explore deep information, ConvE\cite{dettmers2018convolutional} propose a simple multi-layer convolutional architecture for link prediction. ConvE splices and reshapes the subject and relation embedding vectors, then performs a 2D convolution operation, vector flattening, fully connected layer, finally, it matches with all candidate object embeddings. To obtain deeper features, the 3-column matrix of the triple embedding vector is used in ConvKB\cite{nguyen2017novel} and CapsE\cite{vu2019capsule}, ConvKB changes the form of input data of ConvE while CapsE is based on the capsule network. InteractE\cite{vashishth2020interacte} is improved in the convolution step of ConvE, which captures entity and relation feature interactions through three ideas: Feature Permutation, Checkered Reshaping, and Circular Convolution. HypER\cite{balavzevic2019hypernetwork} propose a hypernetwork architecture that generates simplified relation-specific convolutional filters. 

Recently, the approaches in geometric rotation with complex-valued and quaternion-valued embeddings which link prediction have proposed. RotatE\cite{sun2019rotate} introduce relation-based rotation from subject entity to object entity in complex space, which can leverage the advantages of ComplEx\cite{trouillon2016complex} and DistMult\cite{yang2014embedding} and infer multiple relation models. QuatE\cite{zhang2019quaternion} extends this idea to the quaternion space, and prove that by making rotations on two planes rather than on a single plane (RotatE), which has a high degree of freedom. Currently, Rotate3D\cite{gao2020rotate3d} models the non-commutative composition pattern in three-dimensional space with quaternion representation. BoxE\cite{abboud2020boxe} embeds entities as points, and relations as a set of hyper-rectangles (or boxes), which spatially characterize basic logical properties. HittER\cite{chen2020hitter}, a Hierarchical Transformer model consists of two Transformer blocks, joined to learn entity-relation composition and relational contextualization based on information of entity’s neighborhood.

\section{Architecture design }

KGs are usually expressed as form: ($\boldsymbol{E}$, $\boldsymbol{R}$, $\boldsymbol{O}$), in which $\boldsymbol{E}$ is the set of entities, $\boldsymbol{R}$ is the set of relations and $\boldsymbol{O}$ is the set of fact represented as triplets ($h$,$r$,$t$). Link prediction task aims to utilize observed triples to predict hidden triples. We use lowercase letters $h$, $r$, $t$ to denote subject entities, relations, and object entities, and the corresponding bold letters $\boldsymbol{h}$, $\boldsymbol{r}$,  $\boldsymbol{t}$ denote column embedding vectors. Real-valued space and quaternion space are defined as $\mathbb{R}$ and $\mathbb{H}$, respectively.

\subsection{Preliminaries}
Quaternion algebra\cite{hamilton1844lxxviii} is an expansion of the complex algebra, belongs to the hypercomplex number system. Usually, a quaternion $q=a1+b \mathbf{i} + c \mathbf{j} + d \mathbf{k} \in \mathbb{H}$ is composed of one real part and three imaginary parts, where $a$, $b$, $c$, $d \in \mathbb{R}$, and 1, $\mathbf{i}$, $\mathbf{j}$, $\mathbf{k}$ are the quaternion unit basis and $ \mathbf{i} ^2= \mathbf{j} ^2 = \mathbf{k} ^2=\mathbf{ijk} = -1 $. Some basic definitions of quaternion are defined as follows (declaring two quaternions: $q_1 = a_1 + b_1 \mathbf{i}+c_1 \mathbf{j}+d_1 \mathbf{k}$ and $q_2 = a_2 + b_2 \mathbf{i}+c_2 \mathbf{j}+d_2 \mathbf{k}$):

\begin{itemize}

\item Quaternion ordered pairs:

\begin{equation}
\mathrm{q}=[a, \mathbf{v}]=[a, b \mathbf{i}+c \mathbf{j}+d \mathbf{k}] \quad \mathbf{v} \in \mathbb{R}^{3}
\end{equation}i.e. $q_1=[a_1, \mathbf{v}_1]$, $q_2=[a_2, \mathbf{v}_2]$. In this representation, we see the similarity between quaternion and complex number.

\item Product of $\mathbf{i}$, $\mathbf{j}$, $\mathbf{k}$:

\begin{equation}
\begin{array}{c}
\mathbf{i} \mathbf{j}=\mathbf{k} \quad \mathbf{j} \mathbf{k}=\mathbf{i} \quad \mathbf{k} \mathbf{i}=\mathbf{j} \\
\mathbf{j i}=-\mathbf{k} \quad \mathbf{k} \mathbf{j}=-\mathbf{i} \quad \mathbf{i} \mathbf{k}=-\mathbf{j}
\end{array}
\end{equation}

\item Quaternion Addition and Subtraction:

\begin{equation}
q_{1} \pm q_{2}=\left[a_{1} \pm a_{2}, \mathbf{v}_{\mathbf{1}} \pm \mathbf{v}_{\mathbf{2}}\right]
\end{equation}

\item Inner Product:

\begin{equation}
q_{1} \cdot q_{2}=\left[a_{1} \cdot a_{2}, \mathbf{v}_{\mathbf{1}} \cdot \mathbf{v}_{\mathbf{2}}\right]=a_{1} \cdot a_{2}+b_{1} \cdot b_{2}+c_{1} \cdot c_{2}+d_{1} \cdot d_{2}
\end{equation}

\item Conjugate $q^*$ of $q$:

\begin{equation}
q^{*}=[a,-\mathbf{v}]=[a,-b \mathbf{i}-c \mathbf{j}-d \mathbf{k}]
\end{equation}

\item Quaternion Normalization $q^ \triangleleft$ of $q$:

\begin{equation}
q^{\triangleleft}=\frac{q}{\sqrt{a^{2}+b^{2}+c^{2}+d^{2}}}
\end{equation}

\item Hamilton Product (Quaternion Multiplication):

\begin{equation}
\begin{aligned}
q_{1} \otimes q_{2}=\left(a_{1} a_{2}-b_{1} b_{2}-c_{1} c_{2}-d_{1} d_{2}\right) \\
+\left(a_{1} b_{2}+b_{1} a_{2}+c_{1} d_{2}-d_{1} c_{2}\right) \mathbf{i} \\
+\left(a_{1} c_{2}-b_{1} d_{2}+c_{1} a_{2}+d_{1} b_{2}\right) \mathbf{j} \\
+\left(a_{1} d_{2}+b_{1} c_{2}-c_{1} b_{2}+d_{1} a_{2}\right) \mathbf{k}
\end{aligned}
\end{equation}

\end{itemize}

\subsection{QuatDE}

Specifically, we represent the entity embedding matrix $Q \in \mathbb{H}^{ |\boldsymbol{E}| \times k}$ and the relation embedding matrix $W \in \mathbb{H}^{ |\boldsymbol{R}| \times k}$ in the quaternion space, where $|\cdot|$ represents the number of elements in set, and $k$ represents the embedding dimension. We use the following formula to calculate score of a triple $(h,r,t)$ with our model:

\begin{equation}
f_{r}(h, t)=S_{r}\left(Q_{h}\right) \otimes W_{r}^{\triangleleft} \cdot S_{r}\left(Q_{t}\right)
\end{equation}in the formula, $Q_h$, $Q_t$ and $W_r^{\triangleleft}$ donated as:

\begin{equation}
\begin{array}{l}
Q_{h}=a_{h}+b_{h} \mathbf{i}+c_{h} \mathbf{j}+d_{h} \mathbf{k} \\
Q_{t}=a_{t}+b_{t} \mathbf{i}+c_{t} \mathbf{j}+d_{t} \mathbf{k} \\
W_{r}^{\triangleleft}=\frac{a_{r}+b_{r} \mathbf{i}+c_{r} \mathbf{j}+d_{r} \mathbf{k}}{\sqrt{a_{r}^{2}+b_{r}^{2}+c_{r}^{2}+d_{r}^{2}}}
\end{array}
\end{equation}Where $Q_h$, $Q_t$ and unit quaternion $W_r^ {\triangleleft} \in \mathbb{H}^k$. Correspondingly, the coefficient $a_h$, $b_h$, $c_h$, $d_h$, $a_t$, $b_t$, $c_t$, $d_t$, $a_r$, $b_r$, $c_r$ and  $d_r \in \mathbb{R}^k.$ Here, $S_r (e)$ is a dynamic mapping function driven by entity ontology $e$ and relation $r$. Symbol $ \otimes $ defines the Hamilton product operation and symbol $\cdot$ defines the inner product operation, respectively.

For our model, dynamic mapping function lies on entity transition matrix $P \in \mathbb{H}^{|\boldsymbol{E}| \times k}$ and relation transition matrix $V \in \mathbb{H}^{|\boldsymbol{E}| \times k}$. We link $h$, $t$, $r$ to vector $P_h^{\triangleleft}$, $P_t^{\triangleleft}$ and $V_r^{\triangleleft} \in \mathbb{H}^k$, and $S_r (Q_h )$ and $S_r (Q_t )$ are represented as follows:

\begin{equation}
\begin{array}{l}
S_{r}\left(Q_{h}\right)=Q_{h} \otimes P_{h}^{\triangleleft} \otimes V_{r}^{\triangleleft} \\
S_{r}\left(Q_{t}\right)=Q_{t} \otimes P_{t}^{\triangleleft} \otimes V_{r}^{\triangleleft}
\end{array}
\end{equation}where $P_h^{\triangleleft}=a_{ph}+b_{ph} \mathbf{i}+c_{ph} \mathbf{j}+d_{ph} \mathbf{k}$, $P_t^{\triangleleft}=a_{pt}+b_{pt} \mathbf{i}+c_{pt} \mathbf{j}+d_{pt} \mathbf{k}$ are normalized entity transfer vectors, and $V_r^{\triangleleft}=a_{vr}+b_{vr} \mathbf{i}+c_{vr} \mathbf{j}+d_{vr} \mathbf{k}$ is normalized relation transfer vectors, in which $a_{ph}$, $b_{ph}$, $c_{ph}$, $d_{ph}$, $a_{pt}$, $b_{pt}$, $c_{pt}$, $d_{pt}$, $a_{vr}$, $b_{vr}$, $c_{vr}$, $d_{vr} \in \mathbb{R}^k$. The entity transfer vector $P_e$ can adjust the spatial position of the same entity $e$ when facing different triples, and the relation transfer vector $V_r$ projects same entity to different relation-special representation spaces. The dynamic mapping function $S_r (e)$ combines the embedding vector of the entity, the entity transfer vector and the relation transfer vector via Hamilton product. In this way, QuatDE captures more detailed information, and can fit in all triples in the overall knowledge representation of the knowledge graph.

Formally, the score function of QuatDE can be represented as follows:

\begin{equation}
f_{r}(h, t)=\left[Q_{h} \otimes P_{h}^{\triangleleft} \otimes V_{r}^{\triangleleft}\right] \otimes W_{r}^{\triangleleft} \cdot\left[Q_{t} \otimes P_{t}^{\triangleleft} \otimes V_{r}^{\triangleleft}\right]
\end{equation}

\paragraph{\textbf{Loss Function}}:
QuatDE were trained using Adagrad optimizer, by minimizing the negative log-likelihood of the logistic model with $L^2$ regularization on the parameters $w$ of our model:

\begin{equation}
\begin{array}{c}
\mathcal{L}=\sum_{(h, r, t) \in\left\{O \cup O^{-}\right\}} \log \left(1+\exp \left(-l_{(h, r, t)} f_{r}(h, t)\right)\right)
\\ +\lambda|| w \|^{2} \\
\text { in which, } l_{(h, r, t)}=\left\{\begin{array}{cl}
1 & \text { for }(h, r, t) \in O \\
-1 & \text { for }(h, r, t) \in O^{-}
\end{array}\right.
\end{array}
\end{equation}in our model, the parameters $w$ for $L^2$ norm include the embedding vectors and transfer vectors with rate $\lambda$. $O$ is the set of golden triples, and $O^-$ donates the set of negative triplets. Following opinion of Wang\cite{wang2014knowledge}, adopting a Bernoulli distribution to generate negative triplets.

\subsection{Discussion}

\begin{figure}[h]
    \centering
    \includegraphics[width=0.9\columnwidth]{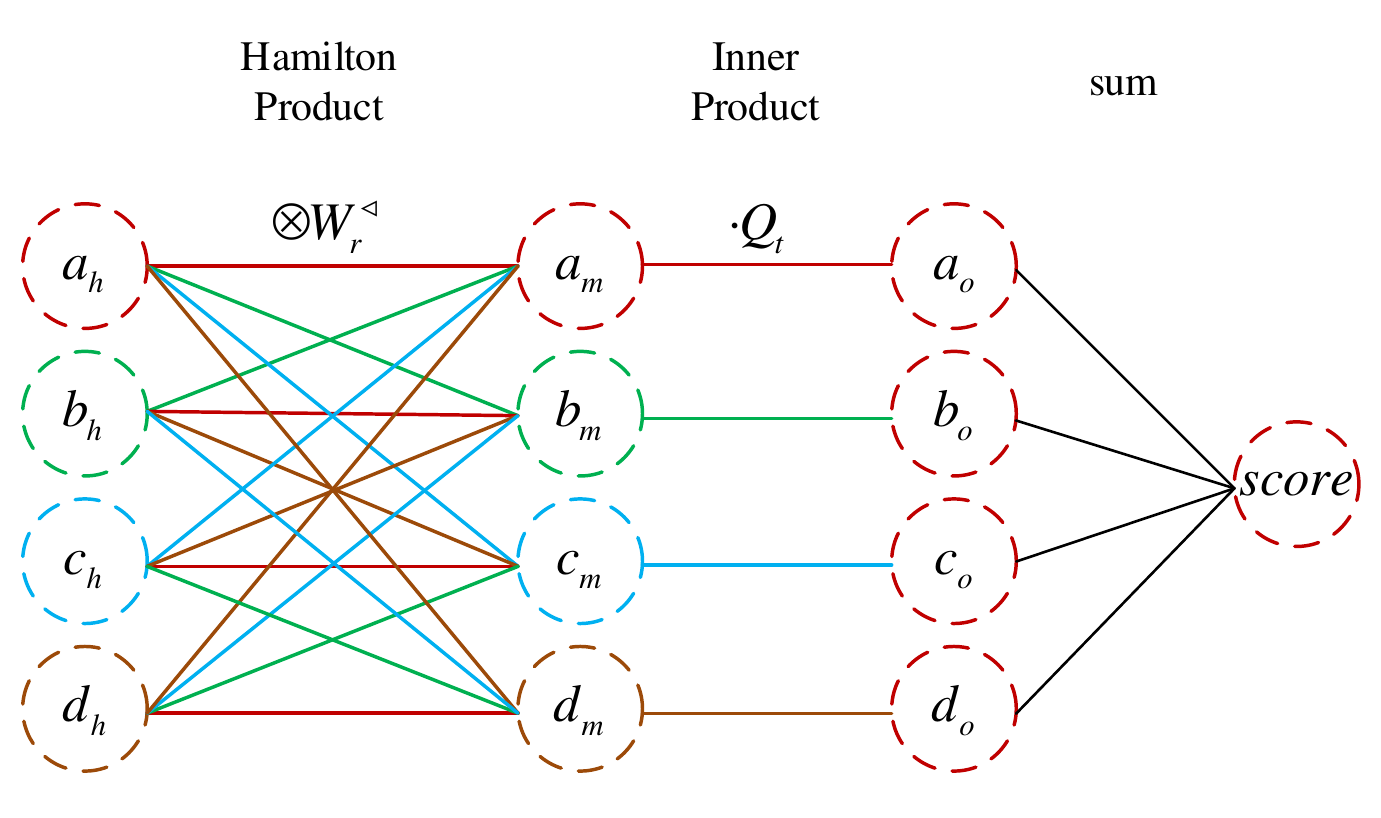}
    \caption{\textbf{architecture of QuatE}}
    \label{img3}
    \end{figure}
    
    \begin{figure*}[h]
    \centering
    \includegraphics[width=0.9\textwidth,height=!]{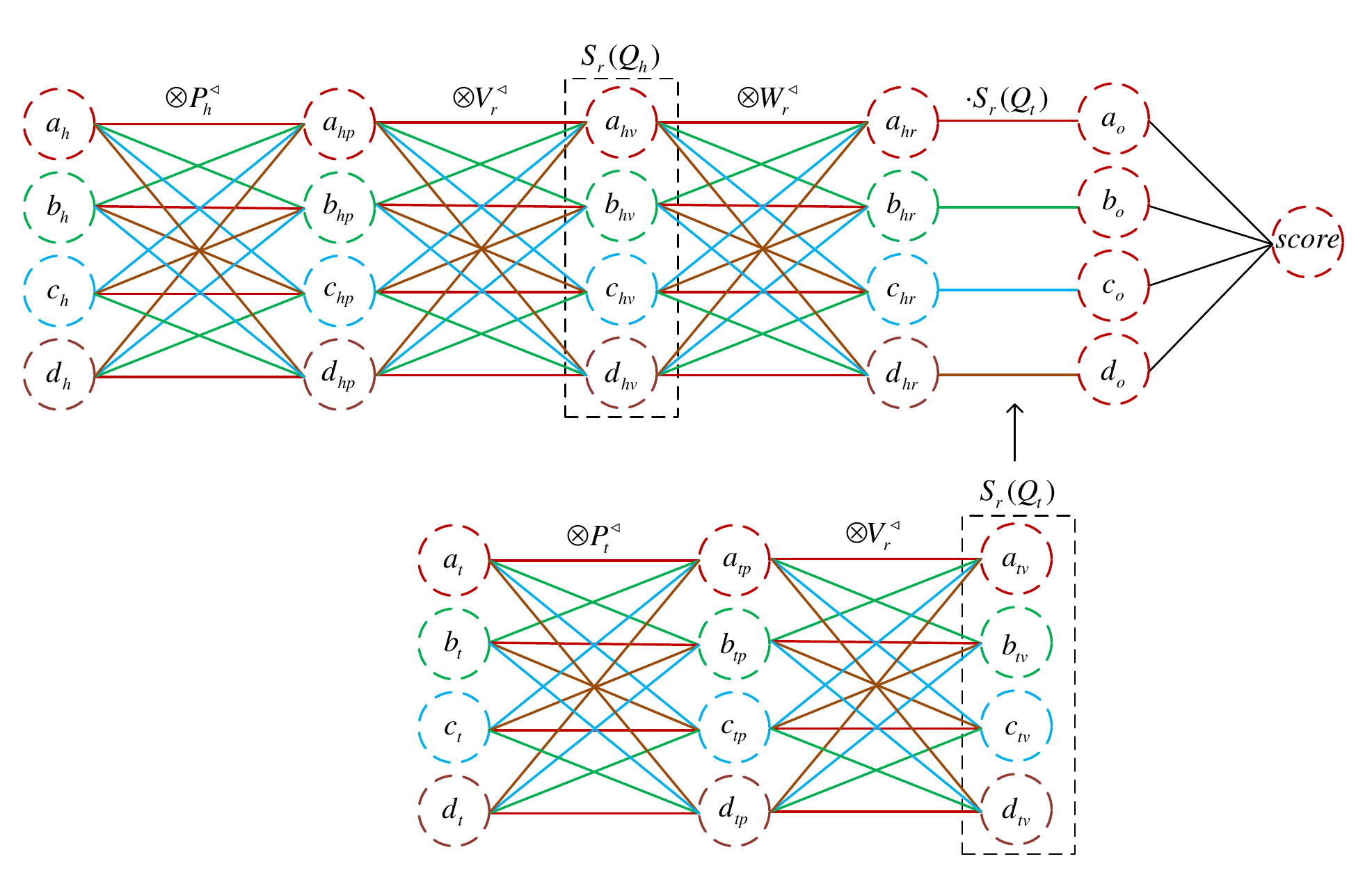}
    \caption{\textbf{architecture of QuatDE}}
    \label{img4}
    \end{figure*}

\paragraph{\textbf{Connection to quaternion-valued neural network (QNN)}}Our work is also inspired by the widespread success of Quaternion number across countless fields, such as heterogeneous image processing\cite{parcollet2019quaternion}, theme identification of telephone conversations\cite{parcollet2017quaternion}, automatic speech recognition\cite{parcollet2018quaternion}. As far as we know, we are the first to use the idea of QNN to connect and explain the knowledge graph embedding models. In this section, we will visualize the model architecture of QuatE (Figure \ref{img3}) and QuatDE (Figure \ref{img4}) from the perspective of QNN, which is not mentioned in QuatE.

As shown in Figure \ref{img3}, the four components of a quaternion are represented by different colors, the weights are also represented with quaternions and transformation is performed with Hamilton product or inner product. In the QuatE model, the input is the quaternion embedding of head; the weights of the first layer are unit quaternion of relation; we can get the intermediate vector via Hamilton product; the intermediate vector is filled into the second layer, and carries out inner product with tail quaternion embedding; Finally, the resulting output vector can be viewed as real number, so we can calculate the triplet score by addition. 

As shown in Figure \ref{img4}, on the basis of QuatE, we add four quaternion feed-forward layers, two layers are used to construct the dynamic strategy function of the head, and the other two are used for that of the tail. The weights of four new layers are closely related to elements of the triplet, i.e. subject transfer vector $P_h^{\triangleleft}$, object transfer vector $P_t^{\triangleleft}$ and relation transfer vector $V_r^{\triangleleft}$, rather than random parameters like traditional neural networks. Meanwhile, increased layers enable more complicated interactions and is less likely to cause over-fitting.

\paragraph{\textbf{Ability to handle complex relations}}We take relation \textbf{\textit{director}} (1-to-N) as an example to describe our solution strategy, and choose the triples (\textbf{Steven Spielberg}, \textbf{\textit{director}}, \textbf{1941}) and (\textbf{Steven Spielberg}, \textbf{\textit{director}}, \textbf{Schindler’s List}), whose labels are true. So according to the score function of QuatDE, $S_{ \text{director}} (Q_{\text{1941}}) \approx S_{ \text{director}} (Q_{ \text{Schindler’s List}})$, but the similarity between $Q_{ \text{1941}}$ and $Q_{\text{Schindler’s List}}$ are still depend on  $P_{\text{1941}}$ and $P_{\text{Schindler’s List}}$ which combine the feature extracted from the triples which include \textbf{1941} or \textbf{Schindler’s List}. QuatDE makes a good trade-off between model efficiency and parameter complexity. Although introducing additional transfer vectors, QuatDE shows strong capabilities in some indicators by setting a smaller embedding dimension.

\paragraph{\textbf{Connection to QuatE}}We applied our motivation to QuatE, and the result proves the feasibility of QuatDE. We believe that through the Hamilton product, the diversity problem of other quaternion models in the knowledge graph completion task can be solved. The transfer vector is removed, QuatDE degenerates to QuatE.

\paragraph{\textbf{Connection to TransH, TransR, TransD}}Although our model belongs to semantic matching models, it also incorporates the advantages of translational distance models. TransH, TransR, and TransD alleviate the problem that TransE does not do well in dealing with 1-to-N, N-to-1, N-to-N relations at different perspectives. TransH and TransR dynamically model the connection structure characteristics between different triples facing different relations. However, TransD do not only consider the diversity of relationships, but also pays attention to entities, which is in line with our paper. The biggest motivation of QuatDE comes from the problem of entities and relations diversity in quaternion space, and we extend this idea to quaternion space via Hamilton product.

\begin{table*}[ht]
    \centering
    \resizebox{\textwidth}{!}{
    \begin{tabular}{@{}l|ccccc|ccccc|ccccc@{}}
    \toprule
    \multirow{2}{*}{Model}               & \multicolumn{5}{c}{WN18RR}           & \multicolumn{5}{c}{FB15K-237}        & \multicolumn{5}{c}{WN18}       \\ \cline{2-16} 
                       & MR              & MRR   & Hit@10 & Hit@3 & Hit@1 & MR  & MRR   & Hit@10 & Hit@3 & Hit@1 & MR  & MRR   & Hit@10 & Hit@3 & Hit@1 \\ \hline
    \textbf{TransE}    & 3384            & 0.226 & 50.1   & -     & -     & 357 & 0.294 & 46.5   & -     & -     & -   & 0.496 & 94.3   & 88.8  & 11.3  \\
    \textbf{ComplEx}   & 5261            & 0.44  & 51.0   & 46.0  & 41.0  & 339 & 0.247 & 42.8   & 27.5  & 15.8  & -   & 0.941 & 94.7   & 94.5  & 93.6  \\
    \textbf{ConvE}     & 4187            & 0.430 & 52.0   & 44.0  & 40.0  & 244 & 0.325 & 50.1   & 35.6  & 23.7  & 374 & 0.943 & 95.6   & 94.6  & 93.5  \\
    \textbf{InteractE} & 5202            & 0.463 & 52.8   & -     & 43.0  & 172 & 0.354 & 53.5   & -     & {\ul 26.3}  & -   & -     & -      & -     & -     \\
    \textbf{rotatE}    & 3340            & 0.476 & 57.1   & 49.2  & 42.8  & 177 & 0.338 & 53.3   & 37.5  & 24.1  & 309 & 0.949 & {\ul 95.9}   & 95.2  & {\ul 94.4}  \\
    \textbf{ATTH}      & -               & 0.486 & 57.3   & 49.9  & \textbf{44.3}  & -   & 0.348 & 54.0   & 38.4  & 25.2  & -   & -     & -      & -     & -     \\
    \textbf{CompGCN}   & 3533            & 0.479 & 54.6   & 49.3  & \textbf{44.3}  & 197 & {\ul 0.355} & 53.5   & {\ul 39.0}  & 25.4  & -   & -     & -      & -     & -     \\
    \textbf{Rotate3D}  & 3328            & \textbf{0.489} & 57.9   & 50.5  & {\ul 44.2}  & 165 & 0.347 & 54.3   & 38.5  & 25.0  & 214 & \textbf{0.951} & \textbf{96.1}   & {\ul 95.3}  & \textbf{94.5}  \\
    \textbf{QuatE}     & {\ul 2314}      & {\ul 0.488} & {\ul 58.2}   & {\ul 50.8}  & 43.8  & \textbf{87}  & 0.348 & {\ul 55.0}   & 38.2  & 24.8  & {\ul 162} & {\ul 0.950} & {\ul 95.9}   & \textbf{95.4}  & \textbf{94.5}  \\ \hline
    \textbf{QuatDE}    & \textbf{1977}   & \textbf{0.489} & \textbf{58.6}   & \textbf{50.9}  & 43.8  & {\ul 90}  & \textbf{0.365} & \textbf{56.3}   & \textbf{40.0}  & \textbf{26.8}  & \textbf{120} & {\ul 0.950} & \textbf{96.1}   & \textbf{95.4}  & {\ul 94.4}  \\ \hline
    \end{tabular}}
    \caption{\textbf{Link prediction result on WN18RR, FB15K-237 and WN18. The best score is in bold, while the second best score is in underline.}}
    \label{tab2}
    \end{table*}

\section{Experiment and Analysis}

\subsection{Datasets}
We evaluate the performance of our model on three general data sets: WN18, WN18RR and FB15K-237. Statistics about the data set are shown in the table \ref{tab1}. FB15K\cite{bordes2013translating} is a subset of Freebase, which is a large dataset contains the facts about sports, actors, movies and others. FB15K-237\cite{toutanova2015observed} was extracted from FB15K and removed inverse relations, which prevent the leakage issue of test triples. WN18\cite{bordes2013translating} is the subset of Wordnet, and it is full of lexical relations between words. WN18 also has many inverse relations, hence, WN18RR\cite{dettmers2018convolutional} is removed inverse relations.

\begin{table}[h]
\centering
\resizebox{.95\columnwidth}{!}{
\begin{tabular}{@{}c|ccccc@{}}
\toprule
\textbf{Dataset}   & \textbf{$|\mathbf{E}|$} & \textbf{$|\mathbf{R}|$} & \textbf{\#training} & \textbf{\#validation} & \textbf{\#test} \\ \hline
\textbf{FB15K-237} & 14541 & 237 & 272115 & 17535 & 20466  \\
\textbf{WN18}      & 40943 & 18  & 141442 & 5000  & 5000   \\
\textbf{WN18RR}    & 40943 & 11  & 86835  & 3034  & 3134   \\ \hline
\end{tabular}}
\caption{\textbf{statistics about the experimental datasets.}}
\label{tab1}
\end{table}

\subsection{Evaluation protocol}

We evaluate related methods on two tasks: link prediction and triplet classification. Link prediction task aims to infer the answer of the query $(h,r,?)$ or $(?,t,r)$ where $?$ means the missing element. So, for each test triple, we calculate the score of all possible triples which can be obtained by substituting subject and object, and sort all scores in descending order.

Mean Rank (MR) and Hit at n are standard evaluation measures for these datasets which are applied in our models. MR measures the average rank of each triplet to predict the correct answer. MRR is defined as the average value of the reciprocated rank, and Hit@n calculates the probability of including the correct entity in the top n ranks. Note that we use the filtered metrics following bordes\cite{bordes2013translating}. The metrics remove all golden triples that appear in either training, validation or test set from the ranking.

\subsection{Training details}

Our code is based on the OpenKE framework and QuatE, and was implemented with PyTorch. We set 100 batches for all datasets. We train our model for 3000 epochs and valid the performance every 300 epochs on three datasets. The dimensionality of embeddings $k \in \{50,100,200,300,400\}$, the number of negative triples for each triple $ \in \{1,5,10\}$, learning rate is selected in $\{0.05,0.1\}$, and $L^2$ regularization parameters $\lambda \in \{0.05,0.1\}$. 

\subsection{Experimental results}

\paragraph{\textbf{Link prediction}}Table \ref{tab2} shows link prediction performance of various models: including translational distance models: TransE\cite{bordes2013translating}, semantic matching models: ComplEx\cite{trouillon2016complex}, ConvE\cite{dettmers2018convolutional}, rotatE\cite{sun2019rotate}, QuatE\cite{zhang2019quaternion}, and recent well-performing models: InteractE\cite{vashishth2020interacte}, ATTH\cite{chami2020low}, CompGCN\cite{vashishth2019composition}, Rotate3D\cite{gao2020rotate3d}. Most of the experimental data are quoted from the original papers. Table \ref{tab2} shows that QuatDE achieved competitive scores than others.

More deeply, we can observe that: 1) QuatDE obtains best scores for MR, MRR, Hit@10 and Hit@3 on WN18RR, and MRR, Hit@10, Hit@3 and Hit@1 on FB15K-237. 2) On WN18, Rotate3D, QuatE and QuatDE are modeled in quaternion space and achieve comparable scores, but our model performs best on MR than the other two. 3) QuatDE is fruitful on FB15K-237, and gains a $0.017$ higher MRR, $1.3\%$ higher Hit@10, $1.8\%$ higher Hit@3 and $2.0\%$ higher Hit@1 than baseline QuatE. 4) The progressive of MR is most obvious. Notably, QuatDE obtains significant improvement of $2314-1977=337$ in MR (which is about $15\%$ relative improvement) on WN18RR, and $162-120=42$ (which is about $26\%$ relative improvement) on WN18.

\paragraph{\textbf{Triplets Classification}}This task aims to judge whether a given triple $(h,r,t)$ is correct or not. Table \ref{tab3} presents triple classification accuracy of different models on WN18RR, FB15K-237 and WN18. We reproduced the code of QuatE and recorded the result of the triple classification task in a table, while others(TransE\cite{bordes2013translating}, TransH\cite{wang2014knowledge}, HolE\cite{nickel2016holographic}, ConvE\cite{dettmers2018convolutional}, ConvKB\cite{nguyen2017novel}, PConvKB\cite{jia2020improving}) are taken from PConvKB\cite{jia2020improving}. Overall, our model QuatDE obtained the best results on three data sets. Especially on FB15K-237 where QuatDE gains considerable improvements of $83.0-81.8=1.2$ compared with QuatE, and $83.0-82.1=0.9$ improvement with PConvKB.

\begin{table}[h]
\centering
\resizebox{.95\columnwidth}{!}{
\begin{tabular}{@{}c|ccc@{}}
\toprule
Model            & WN18RR & FB15K-237 & WN18 \\ \hline
\textbf{TransE}  & 74.0   & 75.6      & 87.6 \\
\textbf{TransH}  & 77.0   & 77.0      & 96.5 \\
\textbf{HoLE}    & 71.4   & 70.3      & 88.1 \\
\textbf{ConvE}   & 78.3   & 78.2      & 95.4 \\
\textbf{ConvKB}  & 79.1   & 80.1      & 96.4 \\
\textbf{PConvKB} & 80.3   & {\ul 82.1}      & 97.6 \\
\textbf{QuatE}   & {\ul 86.7}   & 81.8      & {\ul 97.9} \\ \hline
\textbf{QuatDE}  & \textbf{87.6}   & \textbf{83.0}      & \textbf{98.0} \\ \hline
\end{tabular}}
\caption{\textbf{Triplets classification result.}}
\label{tab3}
\end{table}

\paragraph{\textbf{Multi-relation analysis}}We analyzed the experimental results of complex relations on FB15K-237 and WN18RR. There are 224 relation types in the FB15K-237 test triplets, and QuatDE has achieved an equal or higher score than QuatE on 186 relations (which accounts for $84\%$) when taking Hit@10 as a measure, proving the ability of QuatDE to mitigate the multi-relations. As shown in Table \ref{tab4}, we extract a few examples in each relational pattern (1-to-1, 1-to-N, N-to-1, N-to-N). It’s obviously observed that QuatDE can also increase the prediction accuracy of the 1-to-1 relations, for example, the prediction accuracy of relation \textbf{\textit{campuses}} and \textbf{\textit{educational\_institution}} is $100\%$ in Hit@10, the reason is attributed to intimate feature interaction between the elements of triplet via Hamilton product. And we confirm that QuatDE obtains better MR and Hit@10 than QuatE.

\begin{table}[h]
\centering
\resizebox{0.99\columnwidth}{!}{
\begin{tabular}{@{}c|c|cc@{}}
\toprule
\multicolumn{2}{c}{\multirow{2}{*}{Relation examples}}                                                                                                                         & \multicolumn{2}{c}{QuatE/QuatDE} \\ \cline{3-4} 
\multicolumn{2}{c}{}                                                                                                                                                           & Hit@10         & MR              \\ \hline
\multirow{4}{*}{1-to-1} & /film/film/prequel                                                                                                                                   & 0.75/0.917     & 9.53/3.69       \\ \cline{2-4}
                        & /education/educational\_institution/campuses                                                                                                         & 0.69/1.0       & 25.1/1.0        \\ \cline{2-4}
                        & /location/hud\_county\_place/place                                                                                                                   & 0.81/0.91      & 38.6/11.8       \\ \cline{2-4}
                        & \begin{tabular}[c]{@{}c@{}}/education/educational\_institution\_campus\\        /educational\_institution\end{tabular}                               & 0.61/1.0       & 14.2/1.15       \\ \hline
\multirow{3}{*}{1-to-n} & \begin{tabular}[c]{@{}c@{}}/sports/sports\_league/teams./sports\\       /sports\_league\_participation/team\end{tabular}                             & 0.81/0.91      & 10.6/5.8        \\ \cline{2-4}
                        & \begin{tabular}[c]{@{}c@{}}/education/field\_of\_study/students\_majoring.\\       /education/education/student\end{tabular}                         & 0.31/0.37      & 52.1/41.0       \\ \cline{2-4}
                        & \begin{tabular}[c]{@{}c@{}}/organization/organization/child./organization\\       /organization\_relationship/child\end{tabular}                     & 0.38/0.5       & 28.1/26.4       \\ \hline
\multirow{4}{*}{n-to-1} & \begin{tabular}[c]{@{}c@{}}/film/film/release\_date\_s./film/film\_regional\_release\_date\\       /film\_release\_distribution\_medium\end{tabular} & 0.72/0.78      & 9.3/9.0         \\ \cline{2-4}
                        & /location/location/time\_zones                                                                                                                       & 0.72/0.78      & 43.8/24.8       \\ \cline{2-4}
                        & /film/film/produced\_by                                                                                                                              & 0.42/0.54      & 96.5/90.3       \\ \cline{2-4}
                        & /people/person/nationality                                                                                                                           & 0.55/0.59      & 130.7/118.3     \\ \hline
\multirow{4}{*}{n-to-n} & /location/location/contains                                                                                                                          & 0.47/0.52      & 155.4/117.2     \\ \cline{2-4}
                        & \begin{tabular}[c]{@{}c@{}}/organization/organization\_member/member\_of.\\       /organization/organization\end{tabular}                            & 0.79/0.88      & 11.9/6.7        \\ \cline{2-4}
                        & /film/film/country                                                                                                                                   & 0.51/0.56      & 109.4/91.5      \\ \cline{2-4}
                        & /music/genre/parent\_genre                                                                                                                           & 0.47/0.58      & 27.7/20.6       \\ \hline
\end{tabular}}
\caption{\textbf{Hit@10 and MR of QuatE and QuatDE on some one-to-one, one-to-many, many-to-one, many-to-many relations in FB15K-237.}}
\label{tab4}
\end{table}

Table \ref{tab5} shows  the MRR for each relation on WN18RR, confirming how to effectively model complex information in a space (real number, complex number or quaternion) is a key challenge. A superior-minded model  usually performs better representation capabilities than based-model that only rely on three pure embeddings of the triplet.

\begin{table}[h]
    \centering
    \resizebox{0.99\columnwidth}{!}{
    \begin{tabular}{l | ll}
    Relation Name                 & QuatE & QuatDE \\ \hline
    hypernym                      & 0.173 & \textbf{0.177}  \\
    derivationally\_related\_form & \textbf{0.953} & 0.946  \\
    instance\_hypernym            & \textbf{0.364} & 0.363  \\
    also\_see                     & 0.629 & \textbf{0.640}  \\
    member\_meronym               & 0.232 & \textbf{0.248}  \\
    synset\_domain\_topic\_of     & 0.468 & \textbf{0.493}  \\
    has\_part                     & 0.233 & \textbf{0.239}  \\
    member\_of\_domain\_usage     & \textbf{0.441} & 0.406  \\
    member\_of\_domain\_region    & 0.193 & \textbf{0.294}  \\
    verb\_group                   & \textbf{0.924} & 0.868  \\
    similar\_to                   & \textbf{1.000} & \textbf{1.000}
    \end{tabular}}
    \caption{\textbf{MRR of QuatE vs QuatDE}}
    \label{tab5}
    \end{table}

\begin{figure}[ht]
\centering
\includegraphics[width=0.95\columnwidth,height=!]{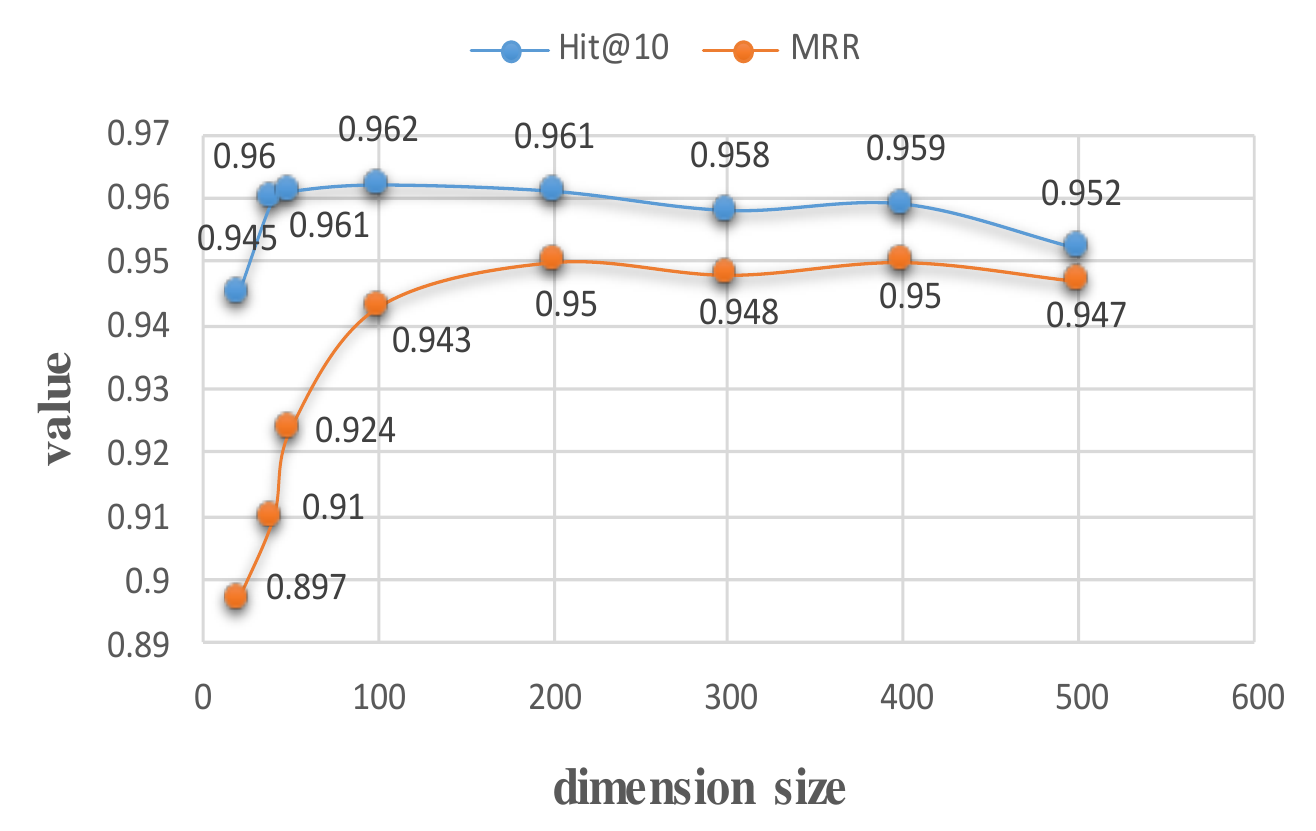}
\caption{\textbf{MRR and Hit@10 with different dimensions of WN18.}}
\label{img5}
\end{figure}

\begin{figure}[ht]
\centering
\includegraphics[width=0.95\columnwidth,height=!]{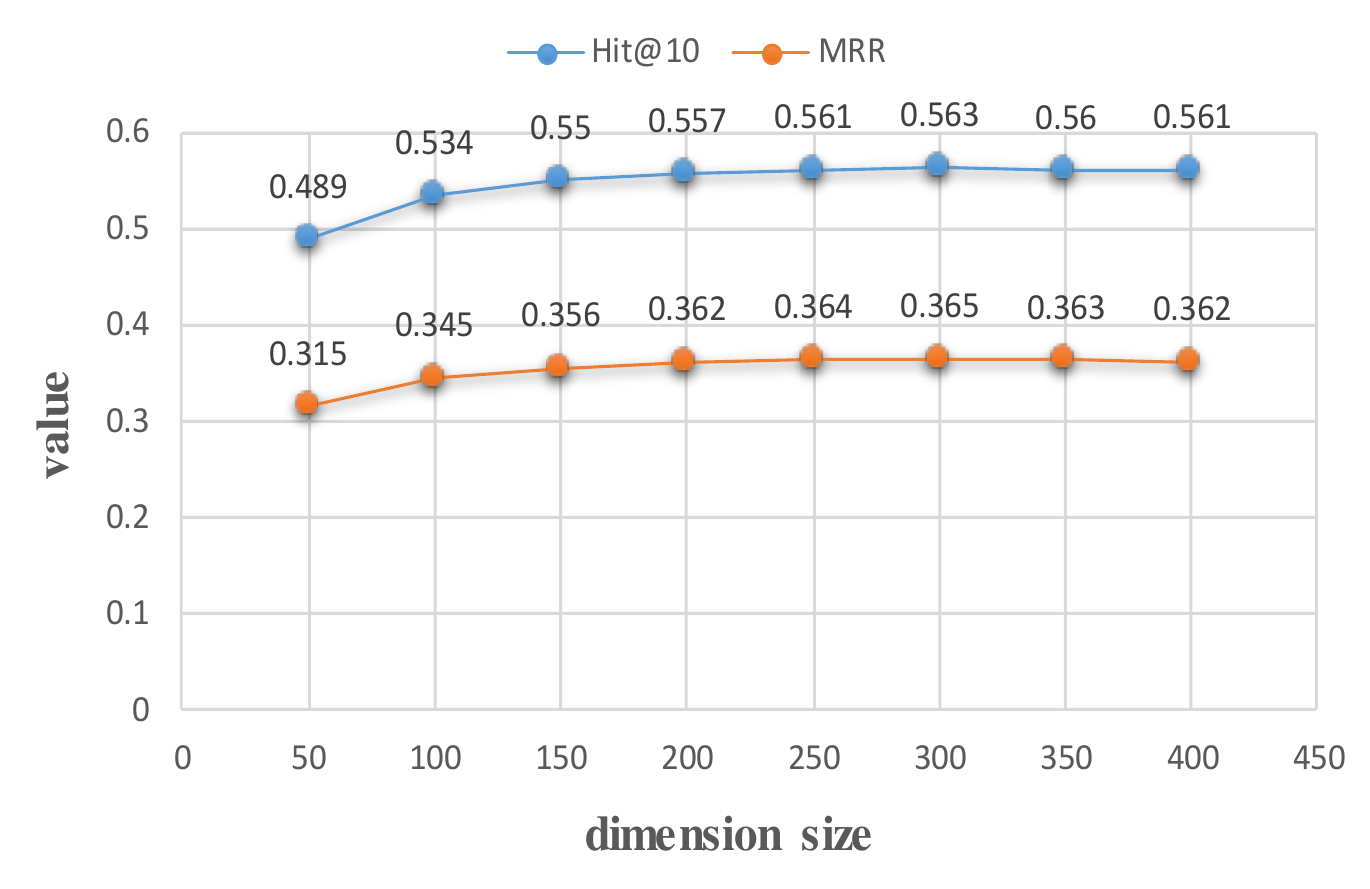}
\caption{\textbf{MRR and Hit@10 with different dimensions of FB15K-237.}}
\label{img6}
\end{figure}

\paragraph{\textbf{Dimension analysis}}We make dimensional analysis on WN18 and FB15K-237, and compared each result when different embedding dimension size K was selected, which is shown in Figure \ref{img5} and Figure \ref{img6}.

Figure \ref{img5} shows the impact of dimension on the QuatDE performance with the changing of dimension size $K$. We can observe that when the dimension is set to $10$, Hit@10 has achieved notable scores higher than $0.94$ and MRR is close to $0.90$. It proves that QuatDE can generate fewer parameters to speed up the model, which allows QuatDE to be extended to a large knowledge graph. Hit@10 is used as the criterion for selecting the best model, and we fix the embedding dimension as $50$, the results of the QuatDE experiment are better than the optimal score ($0.959$) of QuatE whose dimension is selected to $300$ (six times than QuatDE), and QuatDE achieves the optimal model when the dimension is $100$. When the dimension size is selected large than $100$, the curve starts to fluctuate up and down gradually due to the introduction of excessive parameters.

Figure \ref{img6} depicts the dimensional-related experimental results of FB15K-237. On the whole, the resulting curve of FB15K-237 is smoother and more stable than that of WN18, which once again verifies the negative influence of the reverse relations on the experimental results, in detail, WN18 contains a large proportion of reverse relationships, while FB15K-237 excludes that relations. In addition, Hit@10 and MRR are both improved and slowly saturated around $k = 300$.

\section{Conclusion}

In this paper, we propose a novel embedding model QuatDE for link prediction task and triplet classification task in quaternion space. QuatDE utilizes a dynamic mapping method to enhance the character interaction with Hamiltonian product between the triple explicitly, and the model has a higher degree of freedom when training and fitting. The experimental results show that QuatDE outperforms other state-of-the-art models on three benchmark datasets WN18, WN18RR, and FB15K-237. Recently, QDN(quaternion deep network) has been proposed, but the network design of architecture isn’t mature and lacking the interpretability, future work will focus on details and theory of QDN, furtherly explore the application of the quaternion deep network to the knowledge graph completion task, and expand the method to open domain knowledge graph tasks.

\bibliographystyle{aaai}
\bibliography{3019.bibfile}

\end{document}